\documentclass[letterpaper, 10 pt, conference]{ieeeconf}  

\IEEEoverridecommandlockouts                              
\usepackage[utf8]{inputenc}
\usepackage{amsmath,amssymb,stmaryrd,mathtools}
\usepackage{amsfonts}

\usepackage[noend]{algpseudocode}
\usepackage{acronym}
\usepackage{verbatim}
\usepackage{booktabs}
\usepackage{siunitx}
\usepackage{graphics}
\usepackage{graphicx,caption}
\usepackage{suffix}
\usepackage{xstring}
\usepackage{xparse}
\usepackage{expl3}
\usepackage{mathrsfs}
\usepackage{tabularx}
\usepackage{makecell}
\usepackage{array}
\usepackage{hyperref}
\usepackage{cleveref}
\usepackage[ruled,linesnumbered]{algorithm2e}
\usepackage{multirow}
\usepackage{diagbox}
\usepackage{rotating}
\usepackage{bbm}
\usepackage{dsfont}

\usepackage{subcaption}
\usepackage{wrapfig}

\usepackage{tikz}
\usetikzlibrary{arrows,backgrounds,calc}
\usepackage{relsize}
\usepackage{float}
\usepackage{kantlipsum} 
\usepackage{lipsum}
\usepackage{stfloats}
\usepackage{siunitx}

\usepackage[noadjust]{cite}
\usepackage{todonotes}
\usepackage{soul}
\definecolor{smoothgreen}{rgb}{0.7,1,0.7}
\sethlcolor{smoothgreen}

\RequirePackage{luatex85}
\usepackage{pgfplots}
\pgfplotsset{compat=newest}
\pgfplotsset{every axis legend/.append style={%
		cells={anchor=west}}
}
\usepgfplotslibrary{polar}
\usetikzlibrary{arrows}
\tikzset{>=stealth'}

\definecolor{C1}{rgb}{0.0, 0.447, 0.741}
\definecolor{C1_light}{rgb}{0.0, 0.6032388663967612, 1.0}
\definecolor{C2}{rgb}{0.85, 0.325, 0.098}
\definecolor{C3}{rgb}{0.929, 0.694, 0.125}
\definecolor{C4}{rgb}{0.494, 0.184, 0.556}
\definecolor{C5}{rgb}{0.466, 0.674, 0.188}
\definecolor{C6}{rgb}{0.301, 0.745, 0.933}
\definecolor{C7}{rgb}{0.635, 0.078, 0.184}

\usepgfplotslibrary{groupplots}

\usetikzlibrary{shapes.geometric, arrows}

\tikzstyle{startstop} = [rectangle, rounded corners, minimum width=2cm, minimum height=1cm,text centered, draw=black, fill=none]
\tikzstyle{arrow} = [thick,->,>=stealth]

\title{
Intention Aware Robot Crowd Navigation\\with Attention-Based Interaction Graph
}

\author{Shuijing Liu, Peixin Chang, Zhe Huang, Neeloy Chakraborty, Kaiwen Hong, \\ Weihang Liang, D. Livingston McPherson, Junyi Geng, and Katherine Driggs-Campbell
\thanks{S. Liu, P. Chang, Z. Huang, N. Chakraborty K. Hong, W. Liang, D. L. McPherson, and K. Driggs-Campbell are with the Department of  Electrical and Computer Engineering at the University of Illinois at Urbana-Champaign. Emails: \{sliu105, pchang17, zheh4, neeloyc2, kaiwen2, weihang2, dlivm, krdc\}@illinois.edu}
\thanks{J. Geng is with the Department of Aerospace Engineering at Pennsylvania State University. Email: jgeng@psu.edu}
\thanks{This material is based upon work supported by the National Science Foundation under Grant No. 2143435.}
}

\begin{document}
\maketitle
\thispagestyle{empty}
\pagestyle{empty}

\begin{abstract}

We study the problem of safe and intention-aware robot navigation in dense and interactive crowds. 
Most previous reinforcement learning (RL) based methods fail to consider different types of interactions among all agents or ignore the intentions of people, which results in performance degradation.
To learn a safe and efficient robot policy, we propose a novel recurrent graph neural network with attention mechanisms to capture heterogeneous interactions among agents through space and time. 
To encourage longsighted robot behaviors, we infer the intentions of dynamic agents by predicting their future trajectories for several timesteps. 
The predictions are incorporated into a model-free RL framework to prevent the robot from intruding into the intended paths of other agents.
We demonstrate that our method enables the robot to achieve good navigation performance and non-invasiveness in challenging crowd navigation scenarios.
We successfully transfer the policy learned in simulation to a real-world TurtleBot 2i.
Our code and videos are available at {\color{cyan}{\url{https://sites.google.com/view/intention-aware-crowdnav/home}}}.
\end{abstract}

\section{Introduction}
\label{sec:intro}

As robots are increasingly used in human-centric environments, navigation in crowded places with other dynamic agents is an important yet challenging problem.
In crowded spaces, dynamic agents implicitly interact and negotiate with each other. 
The intended goal and preferred walking styles of agents are not observable to the robot, which poses extra difficulties for crowd navigation.

\begin{figure}
\centering
\includegraphics[scale=0.26]{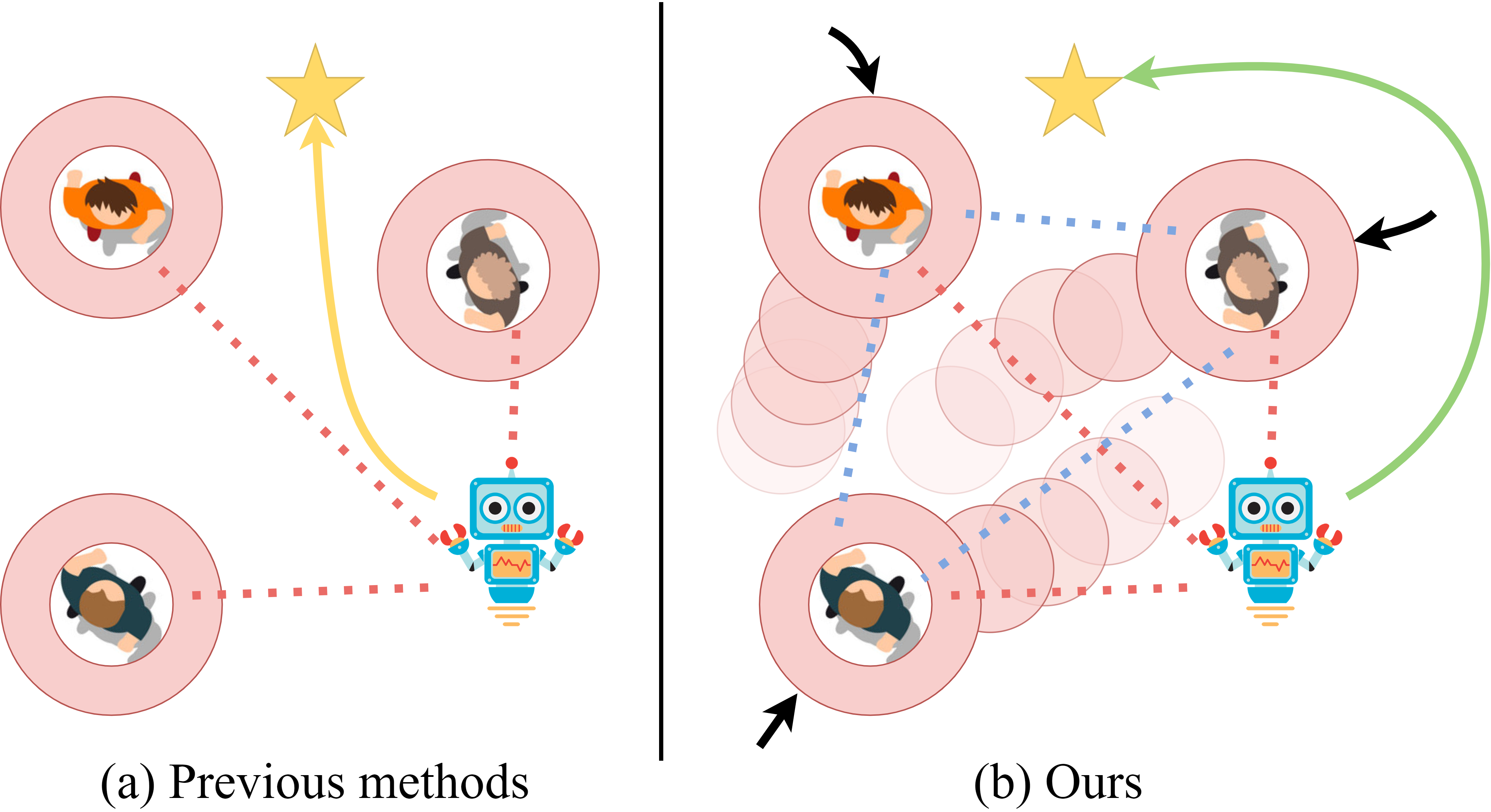}
\caption{\textbf{Previous works and our method.} (a) Previous works consider only robot-human interactions (red-dotted lines) and the safety space of humans are circles centered at the pedestrian's position, which may result in unsafe or shortsighted robot behaviors. (b) Beyond robot-human interactions, our method considers human-human interactions (blue-dotted lines). Our safety space is a set of circles centered at the future positions of humans, which improves the performance and intention awareness of the robot.}
\label{fig:social_zone}
\vspace{-20pt}
\end{figure}

Rising to these challenges, researchers are wrestling with crowd navigation for robots~\cite{fox1997dynamic,van2008reciprocal,chen2019crowd}. 
Reaction-based methods such as Optimal Reciprocal Collision Avoidance (ORCA) and Social Force (SF) use one-step interaction rules to determine the robot's optimal action~\cite{van2011reciprocal, van2008reciprocal,helbing1995social}. 
Learning-based methods model the robot crowd navigation as a Markov Decision Process (MDP) and use neural networks to approximate solutions to the MDP~\cite{chen2017decentralized,chen2017socially,everett2018motion,chen2019crowd,chen2020robot_gaze, liu2020decentralized}. Recently, decentralized structural-RNN (DS-RNN) network is proposed~\cite{liu2020decentralized}. DS-RNN models crowd navigation as a spatio-temporal graph (st-graph) to capture the interactions between the robot and the other agents through both space and time. However, as shown in Fig.~\ref{fig:social_zone}a, the above works only consider the past and current state of humans without explicitly predicting their future trajectories. As a result, the robot sometimes exhibits unsafe or shortsighted behaviors. To deal with this problem, trajectory prediction-based methods plan optimal robot actions conditioned on the predicted future trajectories of other agents~\cite{chen2019relational, li2020socially, katyal2020intent}. However, the prediction-based methods are usually limited to one-step prediction, discrete action space, or a small set of human intentions.

Another weakness of reaction-based and learning-based methods is that they only consider robot-human (RH) interactions but ignore human-human (HH) interactions.
Since the previous works are usually evaluated in sparse crowds or dense crowds with mostly static agents, omitting the human-human interactions does not have a great impact. 
Yet in dense and highly interactive crowds, the performance of these methods degrades due to the lack of human-human interaction modeling. Some attempts have been made to consider HH interactions, but these works either use predefined values to hardcode HH attention scores or do not distinguish the difference between RH and HH interactions~\cite{chen2019relational, chen2020robot_gaze}. Since we have control of the robot but not humans, RH and HH interactions have different effects on the robot's decision-making and thus need to be processed separately.


In this paper, we seek to learn an intention-aware navigation policy which reasons about different interactions in the crowd, as shown in Fig.~\ref{fig:social_zone}b. First, we propose spatio-temporal interaction graph (sti-graph), which captures all interactions in the crowd, and derive a novel neural network from the sti-graph to learn navigation policies. We use two separate multi-head attention networks to address the different effects of RH and HH interactions. The attention networks enable the robot to pay more attention to the important interactions, which ensures good performance when the number of humans increases and the graph becomes complex. 
Second, we enable longsighted robot behaviors by combining the predictions of other agents' intentions with planning. Given any pedestrian trajectory predictor, we design a reward function which encourages the robot to keep away from both current and intended positions of humans, improving both safety and social awareness of the robot.

%

The main contributions of this paper are as follows.
(1) We propose a graph-based policy network that uses attention mechanism to effectively capture the spatial and temporal interactions among heterogeneous agents.
(2) We propose a novel method to incorporate the predicted intentions of other agents into a model-free RL framework. Our method is compatible with any trajectory predictor.
(3) The experiments demonstrate that our method outperforms previous works in terms of navigation performance and social awareness.

\section{Related Works}
\label{sec:related}



\subsection{Reaction-based methods}
\label{sec:related_crowd_nav}


Robot navigation in dynamic crowds has been explored for decades~\cite{fox1997dynamic,hoy2015algorithms, savkin2014seeking,xie2021towards}.
Optimal reciprocal collision avoidance (ORCA) models other agents as velocity obstacles and assumes that agents avoid each other under the reciprocal rule~\cite{van2011reciprocal, van2008reciprocal}. 
Social Force (SF) models the interactions between the robot and other agents using attractive and repulsive forces~\cite{helbing1995social}.
Although ORCA and SF account for mutual interactions among agents, the hyperparameters are sensitive to crowd behaviors and thus need to be tuned carefully to ensure good performance~\cite{long2018towards}. In addition, both methods are prone to failures if the assumptions such as the reciprocal rule no longer hold~\cite{trautman2010unfreezing, liu2020decentralized}.

\subsection{Learning-based methods}
Deep V-Learning addresses the issues of reaction-based methods by learning the interaction rules. Deep V-Learning first uses supervised learning with ORCA as the expert and then uses RL to learn a value function for path planning~\cite{chen2017decentralized,everett2018motion,chen2019crowd,chen2020robot_gaze}. However, Deep V-Learning assumes that state transitions of all agents are known and inherits the same problems as the ORCA expert~\cite{liu2020decentralized}. 
To address these problems, decentralized structural-RNN (DS-RNN) uses a partial graph to model the interactions between the robot and humans~\cite{liu2020decentralized}. Model-free RL is used to train the robot policy without supervised learning or assumptions on state transitions, which achieved better results than ORCA and Deep V-Learning. 
Nevertheless, both Deep V-Learning and DS-RNN only consider the past and current states of humans without explicitly inferring their future intents, which results in shortsighted robot behaviors. 


\subsection{Trajectory prediction-based methods}
\label{sec:related_social_nav}
Utilizing the recent advancements in pedestrian trajectory prediction~\cite{alahi2016social,gupta2018social,vemula2018social,huang2019stgat}, one line of work uses the predicted human states to compute the transition probabilities of MDP and then plans optimal robot actions with search trees~\cite{chen2019relational, eiffert2020path,matsumotomobile}. However, the computation cost of tree search grows exponentially with the size of robot action space. Thus, these methods usually choose discrete and small action spaces, which leads to less natural robot trajectories. 
Another line of work uses the one-step predictions as observations of model-free RL policies~\cite{li2020socially,sathyamoorthy2020densecavoid}. Although the action space can be large and continuous, one-step predictions only capture the instantaneous velocities instead of intentions of pedestrians. Our method makes predictions for several timesteps to capture the long-term intents of pedestrians and modify the reward function based on predictions so that the robot is intention-aware. 
Although some other works attempt to predict the long-term goals of pedestrians to aid planning, the set of all possible goals is finite and small, which limits the generalization of these methods~\cite{katyal2020intent,vemula2017modeling}. In contrast, for our method, the goals of pedestrians can be any position in a continuous two-dimensional space.

\subsection{Attention mechanism for crowd interactions}
\label{sec:related_attn}
Interactions amongst multiple agents contain essential information for both multi-pedestrian trajectory prediction and crowd navigation~\cite{vemula2018social,huang2019stgat, chen2019crowd, liu2020decentralized}.
To capture this information for each agent, self-attention computes attention scores between the agent and all other observed agents~\cite{vaswani2017attention}. This better captures the pair-wise relationships in the crowd than combining the information of other agents with concatenation or an LSTM encoder~\cite{chen2017decentralized,everett2018motion}. 
In crowd navigation of robots and autonomous vehicles, some works use attention to determine the relative importance of each human to the robot~\cite{chen2019crowd, leurent2019social, liu2020decentralized}.  
However, the interactions among humans can influence the robot but are not explicitly modeled.
Other works attempt to model both robot-human (RH) and human-human (HH) attentions using graph convolutional networks~\cite{chen2019relational, chen2020robot_gaze}.
However, Chen and Liu et al. use predefined values to hardcode HH attention scores~\cite{chen2020robot_gaze}, while Chen and Hu et al. do not distinguish the difference between RH and HH interactions~\cite{chen2019relational}. 
As a step further toward this direction, our method recognizes the heterogeneous nature of RH and HH interactions and learns the attention scores separately using different attention networks.

\section{Methodology}
In this section, we first formulate crowd navigation as an RL problem and introduce the prediction reward function. Then, we present our approach to model the crowd navigation scenario as a sti-graph, which leads to the derivation of our network architecture.
\label{sec:methods}

\begin{figure*}[ht]
\centering
\includegraphics[scale=0.55]{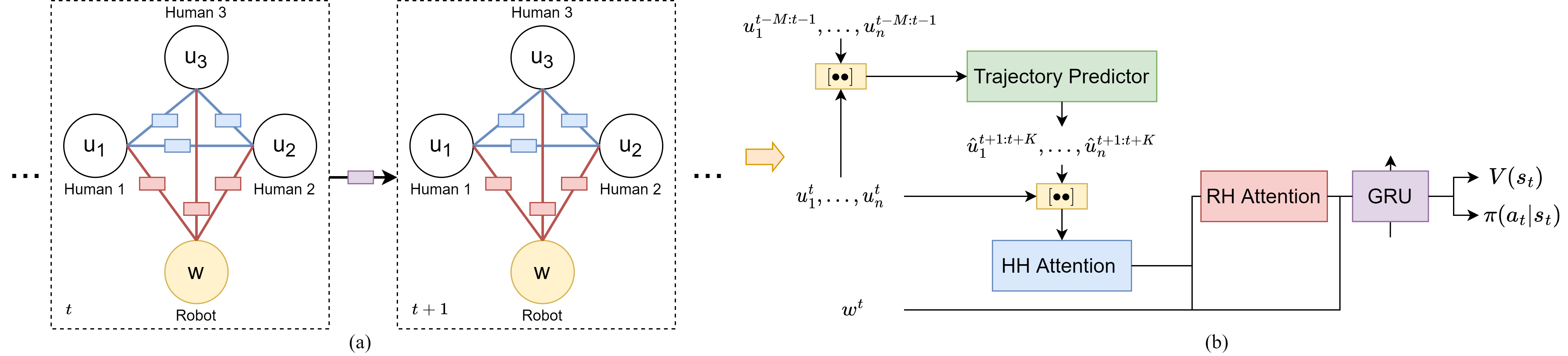}
\caption{\textbf{The spatial-temporal interaction graph and the network architecture.} (a) Graph representation of crowd navigation. The robot node is denoted by $\mathrm{w}$ and the $i$-th human node is denoted by $\mathrm{u}_i$. HH edges and HH functions are in blue, while RH edges and RH functions are in red. Temporal function that connects the graphs at adjacent timesteps is in purple. (b) Our network. A trajectory predictor is used to predict personal zones. Two attention mechanisms are used to model the human-human interactions and robot-human interactions. We use a GRU as the temporal function. }
\label{fig:st}
\vspace{-15pt}
\end{figure*}

\subsection{Preliminaries}
\subsubsection{MDP formulation}
Consider a robot navigating and interacting with humans in a 2D space. We model the scenario as a Markov Decision Process (MDP), defined by the tuple $ \langle \mathcal{S}, \mathcal{A}, \mathcal{P}, R, \gamma, \mathcal{S}_0 \rangle$. 
Let $\mathbf{w}^t$ be the robot state which consists of the robot's position $(p_x, p_y)$, velocity $(v_x, v_y)$, goal position $(g_x, g_y)$, maximum speed $v_{max}$, heading angle $\theta$, and radius of the robot base $\rho$. Let $\mathbf{u}_i^t$ be the current state of the $i$-th human at time $t$, which consists of the human's position $(p_x^i, p_y^i)$. Then, the $K$ predicted future and the $M$ previous positions of the $i$-th human are denoted as $\hat{\mathbf{u}}_i^{t+1:t+K}$ and $\mathbf{u}_i^{t-M:t-1}$, respectively. We define the state $s_t \in \mathcal{S}$ of the MDP as $s_t=[\mathbf{w}^t, \mathbf{u}^t_1, \hat{\mathbf{u}}^{t+1:t+K}_1, ..., \mathbf{u}^t_n, \hat{\mathbf{u}}^{t+1:t+K}_n]$ if a total number of $n$ humans are observed at the timestep $t$, where $n$ may change within a range in different timesteps.


In each episode, the robot begins at an initial state $s_0\in \mathcal{S}_0$. According to its policy $\pi(a_t|s_t)$, the robot takes an action $a_t\in\mathcal{A}$ at each timestep $t$. In return, the robot receives a reward $r_t$ and transits to the next state $s_{t+1}$ according to an unknown state transition $\mathcal{P}(\cdot|s_t, a_t)$. 
Meanwhile, all other humans also take actions according to their policies and move to the next states with unknown state transition probabilities. 
The process continues until the robot reaches its goal, $t$ exceeds the maximum episode length $T$, or the robot collides with any humans. 

\subsubsection{Trajectory predictor}
As shown in Fig.~\ref{fig:st}b, at each timestep $t$, a trajectory predictor takes the trajectories of observed humans from time $t-M$ to $t$ as input and predicts their future trajectories from time $t+1$ to $t+K$:
\begin{equation}
    \hat{\mathbf{u}}^{t+1:t+K}_i = Predictor(\mathbf{u}^{t-M:t}_i), \quad i\in\{1,...,n\}
\end{equation}
We add a circle centered at each predicted position to approximate the intended paths of humans as shown in Fig.~\ref{fig:social_zone}b. If the predictor has learnable parameters, it needs to be pretrained and is only used for inference in our framework. 

\subsubsection{Reward function}
To discourage the robot from intruding into the predicted zones of humans, we use a prediction reward $r_{pred}$ to penalize such intrusions: 
\begin{equation} \label{eq:r_intrusion}
    \begin{split} 
    r^i_{pred}(s_t) &= \min_{k=1,...,K}{ \left(\mathds{1}_i^{t+k} \frac{r_{c}}{2^k}\right)} \\
    r_{pred}(s_t) &= \min_{i=1, ..., n}{r_{pred}^i(s_t)} 
    \end{split}
\end{equation} 
where $\mathds{1}_i^{t+k}$ indicates whether the robot collides with the predicted position of the human $i$ at time $t+k$ and $r_c=-20$ is the penalty for collision. We assign different weights to the intrusions at different prediction timesteps, and thus the robot gets less penalty if it intrudes into the predicted social zone further in the future.

In addition, we add a potential-based reward $r_{pot} = 2(-d_{goal}^t+d_{goal}^{t-1})$ to guide the robot to approach the goal, where $d_{goal}^t$ is the $L2$ distance between the robot position and goal position at time $t$.
Let $S_{goal}$ be the set of goal states, where the robot successfully reaches the goal, and $S_{fail}$ be the set of failure states, where the robot collides with any human.
Then, the whole reward function is defined as 
\begin{equation}
\label{eq:reward}
 \begin{split}
\begin{gathered}
    r(s_t, a_t)  = 
        \begin{cases}
            10, & \text{if } s_t\in S_{goal}\\
            r_c, & \text{if } s_t\in S_{fail}\\
            r_{pot}(s_t)+r_{pred}(s_t), & \text{otherwise}.
        \end{cases}
\end{gathered}
\end{split}
\end{equation}
Intuitively, the robot gets a high reward when it approaches the goal and avoids intruding into the current and future positions of all humans.

The goal of the agent is to maximize the expected return, $R_t=\mathbb{E}[\sum^T_{i=t}\gamma^{i-t}r_{i}]$, where $\gamma$ is a discount factor. The value function $V^\pi (s)$ is defined as the expected return starting from $s$, and successively following policy $\pi$.

\subsection{Spatio-Temporal Interaction Graph}
\label{sec:st_graph}
We formulate the crowd navigation scenario as a sti-graph. 
In Fig.~\ref{fig:st}a, at each timestep $t$, our sti-graph $\mathcal{G}_t = (\mathcal{V}_t, \mathcal{E}_t)$ consists of a set of nodes $\mathcal{V}_t$ and a set of edges $\mathcal{E}_t$. 
The nodes represent the detected agents. 
The edges connect two different detected agents and represent the spatial interactions between the agents at the same timestep. The invisible humans that fall out of the robot's field of view and their corresponding edges are not included in the graph $\mathcal{G}_t$, since the robot cannot predict or integrate their motions in planning.
Since we have control of the robot but not the humans, 
robot-human interactions have direct effects while human-human interactions have indirect effects on the robot deicision-making.
As an example of indirect effects, if human A aggressively forces human B to turn toward the robot's front, the robot has to respond as a result of the interaction between A and B.
Thus, we divide the set of edges $\mathcal{E}_t$ into the human-human (HH) edges that connect two humans and the robot-human (RH) edges that connect the robot and a human.
The two types of edges allow us to factorize the spatial interactions into HH function and RH function. In Fig.~\ref{fig:st}a, the HH and RH functions are denoted by the blue and red boxes respectively and have parameters that need to be learned.
Compared with the previous works that ignore HH edges~\cite{van2011reciprocal,helbing1995social,chen2019crowd,liu2020decentralized}, our method considers the pair-wise interactions among all visible agents and thus scales better in dense and highly interactive crowds.


Since the movements of all agents cause the visibility of each human to change dynamically, the set of nodes $\mathcal{V}_t$ and edges $\mathcal{E}_t$ and the parameters of the interaction functions may change correspondingly.
To this end, we integrate the temporal correlations of the graph $\mathcal{G}_t$ at different timesteps using another function denoted by the purple box in Fig.~\ref{fig:st}a. The temporal function connects the graphs at adjacent timesteps, which overcomes the short-sightedness of reactive methods and enables long-term decision-making of the robot.

To reduce the number of parameters, the same type of edges share the same function parameters. This parameter sharing is important for the scalability of our sti-graph because the number of parameters is kept constant with an increasing number of humans~\cite{jain2016structural}.

\subsection{Network Architecture}
In Fig.~\ref{fig:st}b, we derive our network architecture from the sti-graph. In our network, a trajectory predictor predicts the personal zones of humans. We represent the HH and RH functions as feedforward networks with attention mechanisms, referred to as HH attn and RH attn respectively. We represent the temporal function as a gated recurrent unit (GRU). We use $W$ and $f$ to denote trainable weights and fully connected layers throughout this section.

\subsubsection{Attention mechanisms}
The attention modules assign weights to each edge that connects to an agent, allowing it to pay more attention to important interactions. The HH and RH attentions are similar to the scaled dot-product attention~\cite{vaswani2017attention}, which computes attention score using a query $Q$ and a key $K$, and applies the normalized score to a value $V$. 

\begin{equation} \label{eq:single_head_attn}
    \textrm{Attn}(Q, K, V) = \textrm{softmax}\left(\frac{QK^\top}{\sqrt{d}}\right)V
\end{equation}
where $d$ is the dimension of the queries and keys.

In HH attention, the current states and the predicted future states of humans are concatenated and passed through linear layers to obtain $Q_{HH}^t ,K_{HH}^t, V_{HH}^t\in \mathbb{R}^{n \times d_{HH}}$, where $d_{HH}$ is the attention size for the HH attention.
\begin{equation}
    \begin{split}
        Q_{HH}^t = [\mathbf{u}^{t:t+K}_1, ...,\mathbf{u}^{t:t+K}_n]^\top W_{HH}^{Q} \\
        K_{HH}^t = [\mathbf{u}^{t:t+K}_1, ..., \mathbf{u}^{t:t+K}_n]^\top W_{HH}^{K} \\
        V_{HH}^t = [\mathbf{u}^{t:t+K}_1, ..., \mathbf{u}^{t:t+K}_n]^\top W_{HH}^{V}
    \end{split}
\end{equation}
We obtain the human embeddings $v_{HH}^t\in \mathbb{R}^{n \times d_{HH}}$ using a multi-head scaled dot-product attention, and the number of attention heads is $8$.


In RH attention, $K_{RH}^t\in \mathbb{R}^{1 \times d_{RH}}$ is the linear embedding of the robot states $\mathbf{w}^t$ and $Q_{RH}^t, V_{RH}^t\in \mathbb{R}^{n \times d_{RH}}$ are linear embeddings of the weighted human features from HH attention $v_{HH}^t$. 
\begin{equation}
    \begin{split}
Q_{RH}^t = v_{HH}^t W_{RH}^{Q},\:
K_{RH}^t = \mathbf{w}^t W_{RH}^{K},\:
V_{RH}^t = v_{HH}^t W_{RH}^{V}
 \end{split}
\end{equation}
We compute the attention score from $Q_{RH}^t$, $K_{RH}^t$, and $V_{RH}^t$ to obtain the twice weighted human features $v_{RH}^t\in \mathbb{R}^{1 \times d_{RH}}$ as in Eq.~\ref{eq:single_head_attn}. The number of attention heads is 1.


In HH and RH attention networks, we use binary masks that indicate the visibility of each human to prevent attention to invisible humans. Unlike DS-RNN that fills the invisible humans with dummy values~\cite{liu2020decentralized}, the masks provide unbiased gradients to the attention networks, which stabilizes and accelerates the training.
\subsubsection{GRU}
We embed the robot states $\mathbf{w}^t$ with linear layers $f_{R}$ to obtain $v_R^t$, which are concatenated with the twice weighted human features $v_{RH}^t$ and fed into the GRU:
\begin{equation}
	v_R^t=f_{R}(\mathbf{w}^t), \quad  h^t=\mathrm{GRU}\left(h^{t-1}, ([v_{RH}^t, v_R^t])\right)
\end{equation}
where $h^t$ is the hidden state of GRU at time $t$.
Finally, the $h^t$ is input to a fully connected layer to obtain the value $V(s_t)$ and the policy $\pi(a_t|s_t)$.

\subsubsection{Training}
If the trajectory predictor has learnable parameters, we train the predictor with a dataset of human trajectories collected from our simulator. 
In RL training, we freeze the parameters of the trajectory predictor and only run inference. We train trajectory predictor and RL policy separately because the two tasks have different objectives, and thus joint training is unstable and less efficient.
We use Proximal Policy Optimization (PPO), a model-free policy gradient algorithm, for policy and value function learning~\cite{schulman2017proximal}. To accelerate and stabilize training, we run $16$ parallel environments to collect the robot's experiences. At each policy update, 30 steps of six episodes are used. 


\section{Simulation Experiments}
\label{sec:sim_exp}

In this section, we present our simulation environment, experiment setup, and experimental results in simulation. 
\subsection{Simulation environment}
Our 2D environment simulates a scenario where a robot navigates through a dense crowd in a $12m\times 12m$ space, as shown in Fig.~\ref{fig:qual_result}. Our simulation captures more realistic crowd navigation scenarios than previous works in two aspects~\cite{chen2019crowd, liu2020decentralized}. First, our robot sensor has a limited circular sensor range of $5m$, while the previous works unrealistically assume that the robot has an infinite detection range. Second, the maximum number of humans can reach up to 20, leading to a denser and more interactive human crowd. 

In each episode, the starting and goal positions of the robot and the humans are randomly sampled on the 2D plane. 
To simulate a continuous human flow, humans will move to new random goals immediately after they arrive at their goal positions.
All humans are controlled by ORCA and react only to other humans but not to the robot. This invisible setting prevents our model from
learning an extremely aggressive policy in which the robot
forces all humans to yield while achieving a high reward.

We use holonomic kinematics for simulated robot and humans, whose action at time $t$ consists of the desired velocity along the $x$ and $y$ axis, $a_t=[v_x, v_y]$.  The action space of the robot is continuous with a maximum speed of $1m/s$. We assume that all agents can achieve the desired velocities immediately, and they will keep moving with these velocities for the next $\Delta t$ seconds.
\begin{table}[ht]
  \begin{center}
    \caption{Navigation results with randomized humans}
    \label{tab:rand_results}
    \begin{tabular}{l c c c c c} 
    \toprule
     \textbf{Method} & \textbf{SR}$\uparrow$ & \textbf{NT}$\downarrow$ & \textbf{PL}$\downarrow$ & \textbf{ITR}$\downarrow$ & \textbf{SD}$\uparrow$\\
     
     \midrule
     
     ORCA & 69.0 & 14.77 & 17.67 &19.61 & 0.38 \\
      
      SF & 29.0 & 20.28 & \textbf{15.93} & 17.68 & 0.37 \\
      
      DS-RNN & 64.0 & 16.31 & 19.63 & 23.91 & 0.34 \\
      
      Ours (No pred, HH attn) & 67.0 & 16.82 & 20.19 & 16.13 & 0.37 \\
      
      \midrule
      
      Ours (GST, no HH attn) & 77.0 & \textbf{12.96} & 18.43 & 8.24 & 0.40 \\
      
      Ours (Const vel, HH attn) & 87.0 & 14.03 & 20.14 & 7.00 & 0.42 \\
      
      Ours (GST, HH attn) & \textbf{89.0} & 15.03 & 21.31 & \textbf{4.18} & \textbf{0.44} \\
      
      \midrule
      
      Ours (Oracle, HH attn) & 90.0 & 16.03 & 22.82 & 1.70 & 0.47 \\

      \bottomrule
    \end{tabular}
  \end{center}
  \vspace{-20pt}
\end{table}
We define the update rule for an agent's position $p_x$, $p_y$ as follows:
\begin{equation}
\label{eqn:dynamics}
\begin{split}
    p_x[t+1] &=p_x[t] + v_x[t] \Delta t
    \\
    p_y[t+1] &=p_y[t] + v_y[t] \Delta t  
\end{split}
\end{equation}

We use two simulation environments for experiments. The first environment simulates humans with constantly changing intents and different traits with the following randomizations: First, all humans occasionally change their goals to other random positions within an episode; Second, each human has a random maximum speed $v_{max}\in[0.5, 1.5]m/s$ and radius $\rho\in[0.3, 0.5]m$. In the second environment, all humans have the same maximum speed and radius with no random goal changes. 
The crowds in both environments are dense and interactive, while the randomizations in the first environment pose extra challenges for prediction and decision making.

     
     
     
      


\subsection{Experiment setup}


\subsubsection{Baselines and Ablation Models}
To show the benefits of incorporating predictions, we compare the performance of four models. The first model (No pred, HH attn) does not use any predictor. The second model (Const vel, HH attn) uses the constant velocity predictor which predicts linear future trajectories based on the current velocity of the agent. The third model (GST, HH attn) uses Gumbel Social Transformer (GST) predictor, which is a deep network and can predict non-linear trajectories when the robot has a limited field of view and the tracking of humans is imperfect~\cite{huang2022learning}. The final model (Oracle, HH attn) is an oracle which has the access to the ground truth future trajectory of humans. All predictors predict human trajectories for 5 timesteps. 

We also compare the performance of our model with the representative methods for crowd navigation. 
We choose ORCA~\cite{van2011reciprocal} and SF~\cite{helbing1995social} as the reactive-based baselines and DS-RNN~\cite{liu2020decentralized} as the learning-based baseline. 
To further show the benefits of including human-human interactions in RL policy, the ablation model (GST, no HH attn) uses GST as the trajectory predictor and replaces the HH attention with several fully-connected layers. 

\subsubsection{Training and Evaluation} For RL methods with prediction, we use the reward function as defined in Eq.~\ref{eq:reward}. For RL methods without prediction, the reward function excludes $r_{pred}$. We train all RL methods for $2\times 10^7$ timesteps with a learning rate $ 4\times 10^{-5}$. 

\begin{table}[ht]
  \begin{center}
    \caption{Navigation results without randomized humans}
    \label{tab:no_rand_results}
    \begin{tabular}{l c c c c c}
    \toprule
     \textbf{Method} & \textbf{SR}$\uparrow$ & \textbf{NT}$\downarrow$ & \textbf{PL}$\downarrow$ & \textbf{ITR}$\downarrow$ & \textbf{SD}$\uparrow$\\
     
     \midrule
     
     ORCA & 78.0 & 15.87 & 18.53 & 26.04 & 0.36 \\
      
      SF & 34.0 & 19.95 & \textbf{17.75} & 21.35 & 0.35 \\
      
      DS-RNN & 67.0 & 20.06 & 25.42 & 13.31 & 0.37 \\
      
      Ours (No pred, HH attn) & 82.0 & 19.15 & 22.82 & 14.87 & 0.37 \\
      
      \midrule
      
      Ours (GST, no HH attn) & 82.0 & \textbf{14.21} & 19.35 & 7.22 & 0.40 \\
      
      Ours (Const vel, HH attn) & \textbf{94.0} & 18.26 & 23.98 & 4.49 & \textbf{0.43} \\
      
      Ours (GST, HH attn) & \textbf{94.0} & 17.64 & 22.51 & \textbf{3.06} & \textbf{0.43} \\
      
      \midrule
      
      Ours (Oracle, HH attn) & 94.0 & 15.38 & 21.23 & 2.97 & 0.45 \\

      \bottomrule
    \end{tabular}
  \end{center}
  \vspace{-23pt}
\end{table}
We test all methods with $500$ random unseen test cases. Our metrics include navigation metrics and social metrics. The navigation metrics measure the quality of the navigation and include the percentage success rate (SR), average navigation time (NT) in seconds, and path length (PL) in meters of the successful episodes. The social metrics measure the social awareness of the robot, which include intrusion time ratio (ITR) and social distance during intrusions (SD) in meters. The intrusion time ratio per episode is defined as $c/C$, where $c$ is the number of timesteps that the robot collides with any human’s true future positions from $t+1$ to $t+5$ and $C$ is the length of that episode. The ITR is the average ratio of all testing episodes. We define SD as the average distance between the robot and its closest human when an intrusion occurs. For fair comparison, all intrusions are calculated by ground truth future positions of humans. 


\begin{figure*}[ht]

    \centering
    \includegraphics[width=\linewidth]{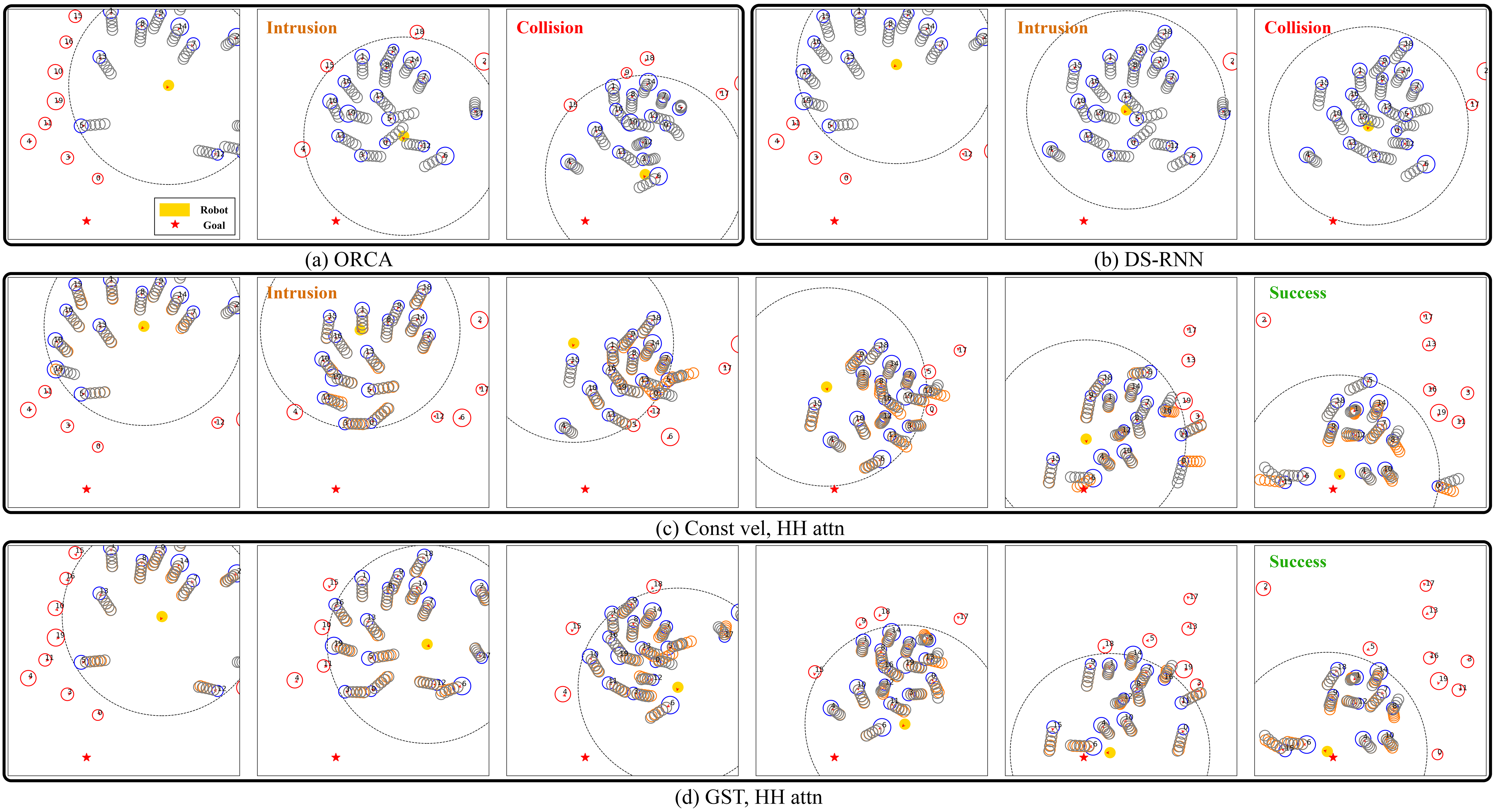}
    
    \vspace{-5pt}
    
    \caption{\textbf{Comparison of different methods in the same testing episode with randomized humans.} The orientation of an agent is indicated by a red arrow, the robot is the yellow disk, and the robot's goal is the red star. We outline the borders of the robot sensor range with dashed lines. Represented as empty circles, the humans in the robot's field of view are blue and those outside are red. The ground truth future trajectories and personal zones are in gray and are only used to visualize intrusions, and the predicted trajectories are in orange. More qualitative results can be found in the video attachment.}
    \vspace{-15pt}
    \label{fig:qual_result}
\end{figure*}

\subsection{Results}
\subsubsection{Effectiveness of HH attention}
The results of robot navigation in two simulation environments are shown in Table~\ref{tab:rand_results} and Table~\ref{tab:no_rand_results}. 
In both tables, for methods with prediction, compared with the ablation model (GST, no HH attn), (GST, HH attn) has $12\%$ higher success rates, lower ITR, and longer SD.
For methods without prediction, (GST, no pred, HH attn) outperforms ORCA, SF, DS-RNN in most cases, especially in the more challenging environment in Table~\ref{tab:rand_results}. Fig~\ref{fig:qual_result} provides an example episode where ORCA and DS-RNN end up with collisions. ORCA assumes that humans take half of the responsibility for collision avoidance, and thus fails immediately if the assumption no longer holds (Fig~\ref{fig:qual_result}a). For DS-RNN, the robot enters instead of deviates from a clutter of humans (Fig~\ref{fig:qual_result}b), indicating that the robot has difficulties reasoning about HH interactions. 
Since (GST, no HH attn) and the aforementioned three baselines only model RH interactions but ignore HH interactions, the HH attention and our graph network contribute to performance gain. 
These results suggest that the HH interactions are essential for dense crowd navigation and are successfully captured by our HH attention mechanism, regardless of the presence of predictor and prediction rewards.

\subsubsection{Intention awareness}
From both tables, we observe that the methods with predictions (last four rows) generally outperform those without predictions (first four rows). 
From the values of (No pred, HH attn), (Const vel, HH attn), and (GST, HH attn), 
we see that the trajectory predictors improve the success rate by around $20\%$ with randomized scenarios (Table~\ref{tab:rand_results}) and by around $12\%$ without randomized scenarios (Table~\ref{tab:no_rand_results}). 
In addition, the models aided by the trajectory predictors have at least $7\%$ lower ITR and $0.02\,m$ larger SD.
The longer NT and PL are caused by our reward function that penalizes the robot to take short yet impolite paths. 
Comparing the two tables, we observe that incorporating predictions gives a larger performance gain in the randomized scenario, where the crowd behaviors are more complicated and thus are more difficult to predict implicitly.
Thus, we conclude that our prediction-aided pipeline enables proactive and long-sighted robot decision making especially in challenging crowd navigation tasks. 


In both scenarios, the performance of (GST, HH attn) is closest to the oracle in terms of all metrics, followed by (Const vel, HH attn).
Since GST can predict non-linear trajectories while constant velocity predictor cannot, this trend also indicates that a more expressive predictor leads to a better policy.
However, we also notice that (GST, HH attn) only outperforms (Const vel, HH attn) with small margins, especially in Table~\ref{tab:no_rand_results}. The reason is that the humans in our simulator are controlled by ORCA, which prefers linear motion if no other agents are nearby. Thus, the best choice of predictors depends on the complexity of crowd behaviors. 

As an illustrative example, with predictions, the robot in Fig~\ref{fig:qual_result}d always keeps a good social distance from the future paths of all humans, while intrusion only happens once in Fig~\ref{fig:qual_result}c.
In contrast, without predictions, the robots in Fig~\ref{fig:qual_result}a and b are notably more aggressive and impolite. Therefore, our prediction-aided RL pipeline effectively prevents the robot from intruding into the intended paths of other agents and thus improves the robot's social awareness.

\section{Real-world Experiments}
\label{sec:real_exp}
To show the effectiveness of our method on real differential drive robots and real crowds, we deploy the model trained in the simulator to a TurtleBot 2i (See \href{https://github.com/Shuijing725/CrowdNav_Sim2Real_Turtlebot}{this link} for implementation details). We use an Intel RealSense tracking camera T265 to obtain the pose of the robot. With an RPLIDAR-A3 laser scanner, we first remove non-human obstacles on a map, and then use a 2D LIDAR people detector~\cite{Jia2020DRSPAAM} to estimate the positions of humans. 
The humans walk at a slow pace to compensate for the robot's maximum speed, which is only $0.5 m/s$.
In a $5m\times 5m$ indoor space with 1 to 4 humans, we run 18 trials and the success rate is $83.33\%$ (Qualitative results are in this \href{https://youtu.be/nxpxhF019VA}{video}). In the failure cases, the turning angle of the robot is too large and it collides with the walls, which indicates that incorporating the environmental constraints is needed in future work. Nevertheless, we demonstrate that our navigation policy interacts with real humans proactively despite noises from the perception module and robot dynamics. 


\section{Conclusion and future work}
\label{sec:conclusion}

We propose a novel pipeline and a network architecture for intention-aware navigation in dense and interactive crowds. 
The experiments show promising results in both simulation and the real world.
Our work suggests that reasoning about past and future spatio-temporal interactions is an essential step toward smooth human-robot interaction. 
However, our work has the following limitations, which open up directions for future work.
(1) The simulated human behaviors are far from realistic. To improve real-world performance, we need a better pedestrian model or a dataset of real pedestrian trajectories for RL training. (2) Our assumption that humans never react to the robot does not hold. Thus, future work needs to incorporate mutual influence among agents.


\textbf{Acknowledgements:} We thank Tianchen Ji and Zhanteng Xie for feedback on paper drafts. We thank members in Human-Centered Autonomy Lab who participated in real-world experiments.



\newpage
\clearpage
\bibliographystyle{IEEEtran}
\bibliography{BibFile}

\begin{thebibliography}{10}
\providecommand{\url}[1]{#1}
\csname url@rmstyle\endcsname
\providecommand{\newblock}{\relax}
\providecommand{\bibinfo}[2]{#2}
\providecommand\BIBentrySTDinterwordspacing{\spaceskip=0pt\relax}
\providecommand\BIBentryALTinterwordstretchfactor{4}
\providecommand\BIBentryALTinterwordspacing{\spaceskip=\fontdimen2\font plus
\BIBentryALTinterwordstretchfactor\fontdimen3\font minus
  \fontdimen4\font\relax}
\providecommand\BIBforeignlanguage[2]{{%
\expandafter\ifx\csname l@#1\endcsname\relax
\typeout{** WARNING: IEEEtran.bst: No hyphenation pattern has been}%
\typeout{** loaded for the language `#1'. Using the pattern for}%
\typeout{** the default language instead.}%
\else
\language=\csname l@#1\endcsname
\fi
#2}}

\bibitem{fox1997dynamic}
D.~Fox, W.~Burgard, and S.~Thrun, ``The dynamic window approach to collision
  avoidance,'' \emph{IEEE Robotics and Automation Magazine}, vol.~4, no.~1, pp.
  23--33, 1997.

\bibitem{van2008reciprocal}
J.~Van Den~Berg, M.~Lin, and D.~Manocha, ``Reciprocal velocity obstacles for
  real-time multi-agent navigation,'' in \emph{IEEE International Conference on
  Robotics and Automation (ICRA)}, 2008, pp. 1928--1935.

\bibitem{chen2019crowd}
C.~Chen, Y.~Liu, S.~Kreiss, and A.~Alahi, ``Crowd-robot interaction:
  Crowd-aware robot navigation with attention-based deep reinforcement
  learning,'' in \emph{IEEE International Conference on Robotics and Automation
  (ICRA)}, 2019, pp. 6015--6022.

\bibitem{van2011reciprocal}
J.~Van Den~Berg, S.~J. Guy, M.~Lin, and D.~Manocha, ``Reciprocal n-body
  collision avoidance,'' in \emph{Robotics research}.\hskip 1em plus 0.5em
  minus 0.4em\relax Springer, 2011, pp. 3--19.

\bibitem{helbing1995social}
D.~Helbing and P.~Molnar, ``Social force model for pedestrian dynamics,''
  \emph{Physical review E}, vol.~51, no.~5, p. 4282, 1995.

\bibitem{chen2017decentralized}
Y.~F. Chen, M.~Liu, M.~Everett, and J.~P. How, ``Decentralized
  non-communicating multiagent collision avoidance with deep reinforcement
  learning,'' in \emph{IEEE International Conference on Robotics and Automation
  (ICRA)}, 2017, pp. 285--292.

\bibitem{chen2017socially}
Y.~F. Chen, M.~Everett, M.~Liu, and J.~P. How, ``Socially aware motion planning
  with deep reinforcement learning,'' in \emph{IEEE/RSJ International
  Conference on Intelligent Robots and Systems (IROS)}, 2017, pp. 1343--1350.

\bibitem{everett2018motion}
M.~Everett, Y.~F. Chen, and J.~P. How, ``Motion planning among dynamic,
  decision-making agents with deep reinforcement learning,'' in \emph{IEEE/RSJ
  International Conference on Intelligent Robots and Systems (IROS)}, 2018, pp.
  3052--3059.

\bibitem{chen2020robot_gaze}
Y.~Chen, C.~Liu, B.~E. Shi, and M.~Liu, ``Robot navigation in crowds by graph
  convolutional networks with attention learned from human gaze,'' \emph{IEEE
  Robotics and Automation Letters}, vol.~5, no.~2, pp. 2754--2761, 2020.

\bibitem{liu2020decentralized}
S.~Liu, P.~Chang, W.~Liang, N.~Chakraborty, and K.~Driggs-Campbell,
  ``Decentralized structural-rnn for robot crowd navigation with deep
  reinforcement learning,'' in \emph{IEEE International Conference on Robotics
  and Automation (ICRA)}, 2021, pp. 3517--3524.

\bibitem{chen2019relational}
C.~Chen, S.~Hu, P.~Nikdel, G.~Mori, and M.~Savva, ``Relational graph learning
  for crowd navigation,'' in \emph{IEEE/RSJ International Conference on
  Intelligent Robots and Systems (IROS)}, 2020, pp. 10\,007 -- 10\,013.

\bibitem{li2020socially}
K.~Li, M.~Shan, K.~Narula, S.~Worrall, and E.~Nebot, ``Socially aware crowd
  navigation with multimodal pedestrian trajectory prediction for autonomous
  vehicles,'' in \emph{IEEE International Conference on Intelligent
  Transportation Systems (ITSC)}, 2020, pp. 1--8.

\bibitem{katyal2020intent}
K.~D. Katyal, G.~D. Hager, and C.-M. Huang, ``Intent-aware pedestrian
  prediction for adaptive crowd navigation,'' in \emph{IEEE International
  Conference on Robotics and Automation (ICRA)}, 2020, pp. 3277--3283.

\bibitem{hoy2015algorithms}
M.~Hoy, A.~S. Matveev, and A.~V. Savkin, ``Algorithms for collision-free
  navigation of mobile robots in complex cluttered environments: a survey,''
  \emph{Robotica}, vol.~33, no.~3, pp. 463--497, 2015.

\bibitem{savkin2014seeking}
A.~V. Savkin and C.~Wang, ``Seeking a path through the crowd: Robot navigation
  in unknown dynamic environments with moving obstacles based on an integrated
  environment representation,'' \emph{Robotics and Autonomous Systems},
  vol.~62, no.~10, pp. 1568--1580, 2014.

\bibitem{xie2021towards}
Z.~Xie, P.~Xin, and P.~Dames, ``Towards safe navigation through crowded dynamic
  environments,'' in \emph{IEEE/RSJ International Conference on Intelligent
  Robots and Systems (IROS)}, 2021, pp. 4934--4940.

\bibitem{long2018towards}
P.~Long, T.~Fan, X.~Liao, W.~Liu, H.~Zhang, and J.~Pan, ``Towards optimally
  decentralized multi-robot collision avoidance via deep reinforcement
  learning,'' in \emph{IEEE International Conference on Robotics and Automation
  (ICRA)}, 2018, pp. 6252--6259.

\bibitem{trautman2010unfreezing}
P.~Trautman and A.~Krause, ``Unfreezing the robot: Navigation in dense,
  interacting crowds,'' in \emph{IEEE/RSJ International Conference on
  Intelligent Robots and Systems (IROS)}, 2010, pp. 797--803.

\bibitem{alahi2016social}
A.~Alahi, K.~Goel, V.~Ramanathan, A.~Robicquet, L.~Fei-Fei, and S.~Savarese,
  ``Social lstm: Human trajectory prediction in crowded spaces,'' in
  \emph{IEEE/CVF Conference on Computer Vision and Pattern Recognition (CVPR)},
  2016, pp. 961--971.

\bibitem{gupta2018social}
A.~Gupta, J.~Johnson, L.~Fei-Fei, S.~Savarese, and A.~Alahi, ``Social gan:
  Socially acceptable trajectories with generative adversarial networks,'' in
  \emph{IEEE/CVF Conference on Computer Vision and Pattern Recognition (CVPR)},
  2018, pp. 2255--2264.

\bibitem{vemula2018social}
A.~Vemula, K.~Muelling, and J.~Oh, ``Social attention: Modeling attention in
  human crowds,'' in \emph{IEEE International Conference on Robotics and
  Automation (ICRA)}, 2018, pp. 1--7.

\bibitem{huang2019stgat}
Y.~Huang, H.~Bi, Z.~Li, T.~Mao, and Z.~Wang, ``Stgat: Modeling spatial-temporal
  interactions for human trajectory prediction,'' in \emph{IEEE International
  Conference on Computer Vision (ICCV)}, 2019.

\bibitem{eiffert2020path}
S.~Eiffert, H.~Kong, N.~Pirmarzdashti, and S.~Sukkarieh, ``Path planning in
  dynamic environments using generative rnns and monte carlo tree search,'' in
  \emph{IEEE International Conference on Robotics and Automation (ICRA)}, 2020,
  pp. 10\,263--10\,269.

\bibitem{matsumotomobile}
K.~Matsumoto, A.~Kawamura, Q.~An, and R.~Kurazume, ``Mobile robot navigation
  using learning-based method based on predictive state representation in a
  dynamic environment,'' in \emph{IEEE/SICE International Symposium on System
  Integration (SII)}, 2022, pp. 499--504.

\bibitem{sathyamoorthy2020densecavoid}
A.~J. Sathyamoorthy, J.~Liang, U.~Patel, T.~Guan, R.~Chandra, and D.~Manocha,
  ``Densecavoid: Real-time navigation in dense crowds using anticipatory
  behaviors,'' in \emph{IEEE International Conference on Robotics and
  Automation (ICRA)}, 2020, pp. 11\,345--11\,352.

\bibitem{vemula2017modeling}
A.~Vemula, K.~Muelling, and J.~Oh, ``Modeling cooperative navigation in dense
  human crowds,'' in \emph{IEEE International Conference on Robotics and
  Automation (ICRA)}, 2017, pp. 1685--1692.

\bibitem{vaswani2017attention}
A.~Vaswani, N.~Shazeer, N.~Parmar, J.~Uszkoreit, L.~Jones, A.~N. Gomez,
  {\L}.~Kaiser, and I.~Polosukhin, ``Attention is all you need,'' in
  \emph{Advances in Neural Information Processing Systems (NeurIPS)}, 2017, pp.
  5998--6008.

\bibitem{leurent2019social}
E.~Leurent and J.~Mercat, ``Social attention for autonomous decision-making in
  dense traffic,'' in \emph{Machine Learning for Autonomous Driving Workshop at
  Advances in Neural Information Processing Systems (NeurIPS)}, 2019.

\bibitem{jain2016structural}
A.~Jain, A.~R. Zamir, S.~Savarese, and A.~Saxena, ``Structural-rnn: Deep
  learning on spatio-temporal graphs,'' in \emph{IEEE/CVF Conference on
  Computer Vision and Pattern Recognition (CVPR)}, 2016, pp. 5308--5317.

\bibitem{schulman2017proximal}
J.~Schulman, F.~Wolski, P.~Dhariwal, A.~Radford, and O.~Klimov, ``Proximal
  policy optimization algorithms,'' \emph{arXiv preprint arXiv:1707.06347},
  2017.

\bibitem{huang2022learning}
Z.~Huang, R.~Li, K.~Shin, and K.~Driggs-Campbell, ``Learning sparse interaction
  graphs of partially detected pedestrians for trajectory prediction,''
  \emph{IEEE Robotics and Automation Letters}, vol.~7, no.~2, pp. 1198--1205,
  2022.

\bibitem{Jia2020DRSPAAM}
D.~Jia, A.~Hermans, and B.~Leibe, ``{DR-SPAAM: A Spatial-Attention and
  Auto-regressive Model for Person Detection in 2D Range Data},'' in
  \emph{IEEE/RSJ International Conference on Intelligent Robots and Systems
  (IROS)}, 2020, pp. 10\,270--10\,277.

\end{thebibliography}
\clearpage
\end{document}